\documentclass[wcp]{jmlr}

\usepackage{booktabs}

\usepackage{xcolor}
\usepackage{colortbl}
\definecolor{lgray}{gray}{0.9}
\definecolor{dgray}{gray}{0.5}
\usepackage{multirow}
\usepackage{rotating}

\usepackage{amsmath,amsfonts,bm}

\def\1{\bm{1}}

\def\rvd{{\mathbf{d}}}
\def\rve{{\mathbf{e}}}

\def\rvh{{\mathbf{h}}}

\def\rvr{{\mathbf{r}}}

\def\rvv{{\mathbf{v}}}

\def\rvx{{\mathbf{x}}}
\def\rvy{{\mathbf{y}}}
\def\rvz{{\mathbf{z}}}

\def\va{{\bm{a}}}
\def\vb{{\bm{b}}}

\def\vx{{\bm{x}}}

\DeclareMathAlphabet{\mathsfit}{\encodingdefault}{\sfdefault}{m}{sl}
\SetMathAlphabet{\mathsfit}{bold}{\encodingdefault}{\sfdefault}{bx}{n}

\def\gE{{\mathcal{E}}}

\def\gG{{\mathcal{G}}}

\def\gN{{\mathcal{N}}}

\def\gS{{\mathcal{S}}}

\def\gV{{\mathcal{V}}}

\def\gZ{{\mathcal{Z}}}

\newcommand{\E}{\mathbb{E}}

\newcommand{\R}{\mathbb{R}}

\newcommand{\gtext}[1]{{\color{dgray} #1}}

\makeatletter
\let\Ginclude@graphics\@org@Ginclude@graphics 
\makeatother

\jmlrvolume{XX}
\jmlryear{2022}
\jmlrworkshop{ACML 2022}

\title[GrannGAN]{Graph annotation generative adversarial networks}

  \author{\Name{Yoann Boget} \Email{yoann.boget@hesge.ch}\\
  \addr University of Geneva and Geneva School for Business administration HES-SO\\
  Rue de la Tambourine 17\\
  Carouge, Switzerland
  \AND
  \Name{Magda Gregorova} \Email{magda.gregorova@fhws.de}\\
  \addr Center for Artificial Intelligence  and Robotics (CAIRO), FHWS\\
  Franz-Horn-Strasse 2\\
  Würzburg-Schweinfurt, Germany
  \AND
  \Name{Alexandros Kalousis} \Email{alexandr.kalousis@hes-so.ch}\\
  \addr Geneva School for Business administration HES-SO \\
  Rue de la Tambourine 17\\
  Carouge, Switzerland
 }

\editors{Emtiyaz Khan and Mehmet Gonen}

\begin{document}

\maketitle

\begin{abstract}
We consider the problem of modelling high-dimensional distributions and generating new 
examples of data with complex relational feature structure coherent with a graph skeleton.
The model we propose tackles the problem of generating the data features constrained by the specific graph structure of each data point by splitting the task into two phases.
In the first it models the distribution of features associated with the nodes of the given graph, in the second it complements the edge features conditionally on the node features.
We follow the strategy of implicit distribution modelling via generative adversarial network (GAN) combined with permutation equivariant message passing architecture operating over the sets of nodes and edges.
This enables generating the feature vectors of all the graph objects in one go (in 2 phases) as opposed to a much slower one-by-one generations of sequential models, prevents the need for expensive graph matching procedures usually needed for likelihood-based generative models, and uses efficiently the network capacity by being insensitive to the particular node ordering in the graph representation.
To the best of our knowledge, this is the first method that models the feature distribution along the graph skeleton allowing for generations of annotated graphs with user specified structures.
Our experiments demonstrate the ability of our model to learn complex structured distributions through quantitative evaluation over three annotated graph datasets.

\end{abstract}

\section{Introduction}\label{sec:intro}

Modern deep learning approaches for learning high-dimensional data distributions and synthesizing new data examples have achieved great successes in a  number of domains. 
We focus here on the particularly challenging problem of generating new data examples $\rvx_i \in \R^{m_i}$ with different dimensionalities $m_i$ between individual instances and complex relational structures in the feature spaces organized in graphs.

More concretely, we focus on learning the distributions and generating new examples of features associated with nodes and edges of annotated graphs conditioned on the graph structure.


We formulate the problem as that of modeling a conditional distribution, where we learn to generate the node and edge features conditionally on the given graph skeleton (non-annotated graph). 
In this respect, our paper complements the existing tool-set of models for non-annotated graph generations (e.g. \cite{you_graphrnn_2018,liao_efficient_2019}).
The non-annotated graph can be generated by one of those or, more interestingly, can be provided by the user based on the needs of a downstream task that requires a specific graph structure (e.g. basis for scaffold-based \emph{de novo} drug discovery or molecular docking, types and interactions of particles in high energy physic, individual features and types of relations in a social network).

The graph annotation generative adversarial networks (GrannGAN) method that we propose follows the strategy of implicit distribution modeling via adversarial training \cite{goodfellow_generative_2014}, allowing us to sample new graphs from a distribution $p$ approximating the true $p^*$ without explicitly formulating the $p$ distribution function.
In principle, the method can generate node of edge features independently. 
In practice, we generate the node and edge features in two phases depicted in Figure \ref{fig:archi}.
In the first phase, our model samples the node features conditioning on the graph skeleton. In the second phase, it uses the generated node features as an additional conditional variable to sample the edge features. 

The entire model architecture relies on the permutation equivariant transformations of message passing neural networks (MPNN) \cite{gilmer_neural_2017}.
Thanks to these, we circumvent the difficulties related to the ordering of graph objects in their representation.
This critical property of the method allows for efficient use of the model capacity, which needs to learn neither a particular heuristic for unique representation ordering (as in the case of linearized representations) nor the complete set of equivalent permutations. 

In the following, we describe the newly proposed GrannGAN method and provide some details of the technical implementation in sections \ref{sec:method}.
We then position the method within the existing state of research and document its performance competitive with the best of the well-established methods for graph generations on a set of experiments in section \ref{sec:related} and \ref{sec:experiments}. 
We conclude with a discussion of possible future directions in section \ref{sec:conclude}.

\section{GrannGAN}\label{sec:method}

Let $\gG = \{ \gV, \gE, V, E \} $ be an undirected graph with a set of nodes (vertices) $\gV$ and a set of edges $\gE$ between pairs of nodes in the graph, $V$ and $E$ are the corresponding node and edge features.
Let $\nu_i$ denote a node $i$ and $\rvv_i \in \R^d$ the features associated with that node. An edge $\epsilon_{ij}$ is connecting the pair of nodes $\nu_i$ and $\nu_j$ and $\rve_{ij} \in \R^c$ are features corresponding to that edge.
We consider only the case where the edges are undirected so that $\epsilon_{ij}= \epsilon_{ji}$ and $\rve_{ij} = \rve_{ji}$, but the model can be easily extended to the case with directed edges. 
We further use the term \emph{skeleton} and the letter $\gS$ to refer to the non-annotated graph $\gS = \{\gV, \gE\}$ corresponding to $\gG$.

\subsection{Model factorization}\label{sec:factorization}

In our approach, we model the underlying graph distribution in the following factorization
\begin{align}\label{eq:factorization}
p(\gG) = p(\gS, V, E) = p(\gS) p(V \rvert \gS) p(E \rvert \gS, V) \enspace .
\end{align}

Our model consists of modeling $p(V \rvert \gS)$ and $p(E \rvert \gS, V)$ with conditional GAN. Note that we could equivalently reverse the conditioning order and model $p(E \rvert \gS)$ and $p(V \rvert \gS, E)$. 
In our experiments, we found the first option yielding better results. So, we keep this ordering for the following. 

During training, we sample the graph skeleton from the data distribution $p^*(\gS)$. At inference, we can either sample the graph skeleton or keep it fixed depending on the task. 

\subsection{Implicit data generation}\label{sec:cGAN}

We use the Wasserstein-GAN (WGAN) \cite{arjovsky_wasserstein_2017} formulation of adversarial training with spectral normalization \cite{miyato_spectral_2018} in all linear layers of the critic to enforce 1-lipschitzness.
To sample new examples of data $\rvx$ from a model distribution $p_\theta(\rvx)$ WGAN uses a generator $g_\theta$ mapping from a random latent variable $\rvz$ to the output $\rvx = g_\theta(\rvx)$.
The generator is learned to minimize the Wasserstein-1 (or Earth-Mover) distance  $W(p_\theta(\rvx), p^*(\rvx))$ between the implicit model distribution and the true generative distribution through a min-max optimization. 

\begin{align}\label{eq:WGAN}
\min_\theta \max_\varphi \ \E_{(y) \sim p^*} f_\varphi(\rvy) - \E_{z \sim p(z)} f_\varphi\big(g_\theta(\rvz)\big) \enspace ,
\end{align}
where $f_\varphi$ is the K-Lipschitz critic function mapping from the data $\rvy$ (real or generated) to a real-valued score $f_\varphi(\rvy) \in \R$.

We use the conditional WGAN formulation twice in the GrannGAN pipeline sketched out in figure \ref{fig:archi}.
Following the factorization in equation \eqref{eq:factorization} we first model the conditional distribution $p(V \rvert \gS)$ in a \emph{node-annotation} phase.
Here, the method generates the node features $\hat{V} = g_{V, \theta}(\gZ, \gS)$ from the set of latent noise variables $\gZ$ and the skeleton $\gS$. So, equation \ref{eq:WGAN} becomes
\begin{align}
    \min_\theta \max_\varphi \ \E_{(V, \gS) \sim p^*} f_{V, \varphi}(V,  \gS) - \E_{z \sim p(z), \gS \sim p^*} f_{V, \varphi}\big(g_{V, \theta}(\gZ, \gS),  \gS \big) \enspace .
\end{align}

Similarly in the following \emph{edge-annotation} phase, we model the conditional $p(E \rvert \gS, V)$.
The edge features $\hat{E} = g_{E, \theta}(\gZ, V, \gS)$ are generated conditionally on the previously generated node features by including these as an additional input variable to the generator and critic. In this phase, equation \ref{eq:WGAN} becomes

\begin{align}
    \min_\theta \max_\varphi \ \E_{(E, V, \gS) \sim p^*} f_{E, \varphi}(E, V, \gS) - \E_{z \sim p(z), (V, \gS) \sim p^*} f_{E, \varphi}\big(g_{E,\theta}(\gZ, V, \gS), V, \gS \big) \enspace ,
\end{align}

Note that the conditioning on the skeleton $\gS = (\gV, \gE)$ acts on the computational graph of the critic $f$ and of the generator $g$, i.e. on the structure of the Message Passing Neural Network (MPNN, explained in detail in the section \ref{sec:MPNN}). 
This special way of conditioning through the computational graph is one of the novelties of our method.


\subsection{Implementation}\label{sec:Implementation}

In practice, we start by sampling the skeleton $\gS = \{\gV, \gE\}$ from the data. 
For all node $\nu_i \in \gV$, we sample a random noise vector $\rvz_i$. 
So, we get $\gZ = \{\rvz_i\}_1^n$ and, by construction $|\gV| = |\gZ|$. 
Doing so, we obtain a graph with random noise vectors as initial latent representation of the node features $\gG_0 = \{\gV, \gE, \gZ\}$ (and without edge feature). 
This graph is the input of the \emph{node-annotation} generator.
Similarly, for the edge generation, we sample a graph from the data without edge feature. 
We also sample $\gZ = \{\rvz_i\}_1^m$, where $m = |\gE|$. 
The input of the \emph{edge-annotation} generator is the graph $\gG_0 = \{\gV, \gE, V, \gZ\}$ 
with latent random noise vectors as initial latent edge representation. 
The generators are permutation-equivarient functions. 
They output node or edge features following the order of the input.

The critics receive alternatively real or generated graphs (without edge feature during the annotation step). The critics are permutation-invariant functions. 
We provide details the generators and the critics in the next subsection.

\begin{figure}[htp]
\centering
\includegraphics[width=0.8\textwidth]{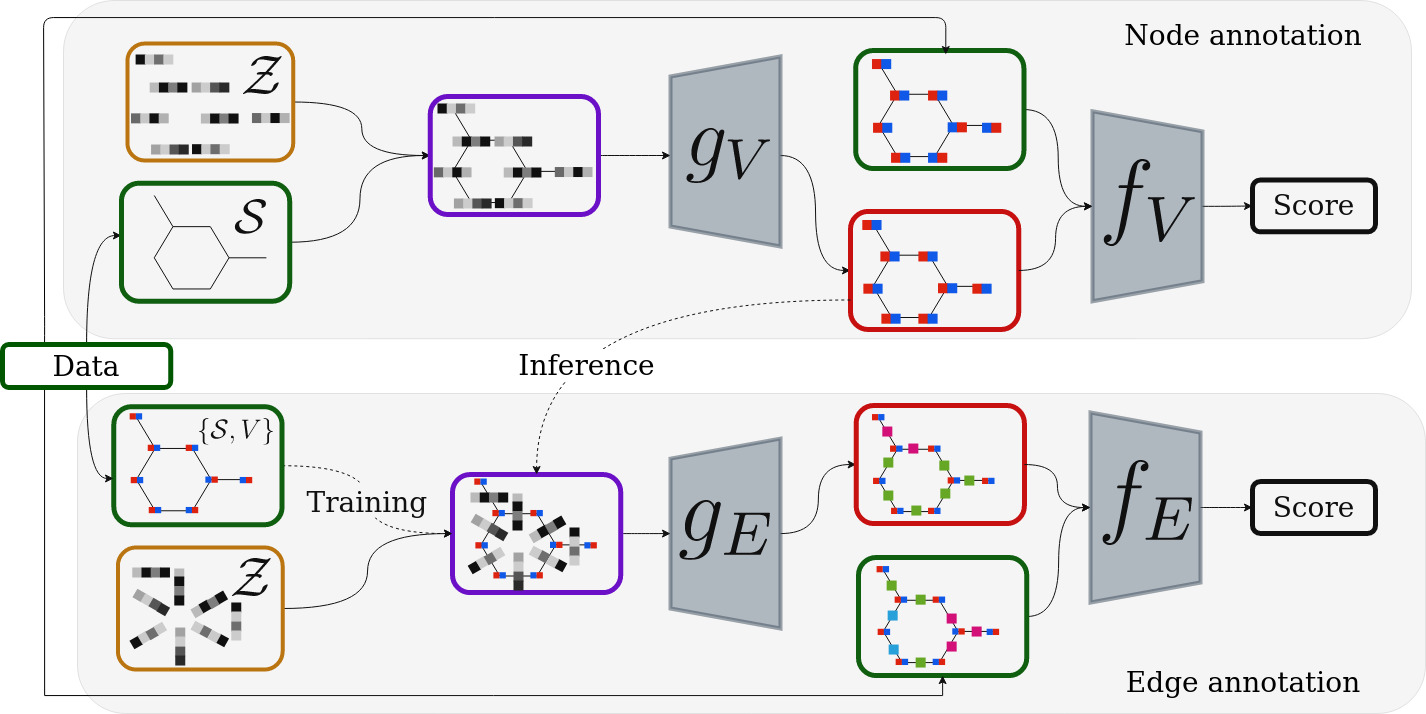}
\caption{GrannGAN architecture: grayscale boxes represent random noise, colored boxes represent node and edge features.}
\label{fig:archi}
\end{figure}

The two steps (the two conditional GANs) are trained independently. 
Using teacher forcing, we use real data for the conditioning during training. 
For inference, we generate the edges features by conditioning on the previously generated node features. 
The Figure \ref{fig:archi} illustrates the model architecture.

The software implementation of our method together with instructions for replicating our experiments are available at \url{https://github.com/yoboget/GrannGAN}.
In this section we provide some details of the implementation to help the reader understand important design decisions.

\subsection{Message passing neural network}\label{sec:MPNN}

In both the node- and the edge-annotation phases described in section \ref{sec:cGAN} all GrannGAN generator $g$ and critic $f$ functions are message passing neural networks (MPNN) \cite{gilmer_neural_2017}.
As mentioned here-above, the node-neighbourhood structure of the MPNNs is derived from the conditioning on the graph skeleton $\{\gV \gE\}$. Therefore, the skeleton acts on the $i$ and $j$ indices. 

We use the following equations to perform $L$ update steps over the hidden states $\rvh_{i}^{(l)}$ and $\rvr_{ij}^{(l)}$ of each node and edge in the graph respectively 
\begin{align}
\rvr_{ij}^{(l+1)} & = \phi_{r}^{(l)}([\rvh_{i}^{(l)}, \rvh_{j}^{(l)}, \rvr_{ij}^{(l)}]) \label{eq:edge} \\
\widetilde{\rvh}_{i}^{(l+1)} & = \sum_{j \in \gN_i} \frac{1}{\sqrt{d_i}{\sqrt{d_j}}} \rvr_{ij}^{(l+1)} \label{eq:nodeaggreg} \\
\rvh_{i}^{(l+1)} & = \phi_h^{(l)}([\rvh_{i}^{(l)}, \widetilde{\rvh}_{i}^{(l+1)}]) \label{eq:node} \enspace .
\end{align}
Here $\phi$ are learned differentiable functions (we use small feedforward networks), $\gN_i$ is the first order neighbourhood of node $\rvv_i$ as given by the graph skeleton $\gS$, $d_i$ is its degree, and $[\va,\vb]$ is the concatenation of vectors $\va$ and $\vb$.

We outline the whole generative pipeline in figure \ref{fig:archi}.
In the generator of the node-annotation phase $g_{\theta_\gV}$ we initiate the node hidden states with a latent random noise vectors $\rvh_i^{(0)} = \rvz_i, \, \forall \rvv_i \in \gV$ (and drop $\rvr_{ij}^{(0)}$ from inputs to $\phi_r^{(0)}$).
The generated node features are the hidden states of the last update step $\widehat{\rvv}_i = \rvh_{i}^{(L)}$.
In contrast, in the edge-annotation generator $g_{\theta_\gE}$ the latent random noise is used for initiating the edge hidden states $\rvr_{ij}^{(0)} = \rvz_{ij}, \, \forall \rve_{ij} \in \gE$ while the node hidden states are the node features generated from the previous phase $\rvh_{i}^{(0)} = \widehat{\rvv}_i$.
The generated edge features are the hidden edge states of the last update step $\widehat\rve_{ij} = \rvr_{ij}^{(L)}$.

The critic functions MPNNs are initiated from the real data examples as $\rvh_i^{(0)} = \rvv_i$ and $\rvr_{ij}^{(0)} = \rve_{ij}$ or from the synthetic examples of the generators respectively.
To produce the critic scores, the last-step update functions ($\phi_h^{(L)}$ for the node critic and $\phi_r^{(L)}$ for the edge critic) have scalar outputs that are averaged to enter the loss in equation \eqref{eq:WGAN}. Therefore, the critic evaluates each node, aggregating information from all the nodes included in a radius of $2(L-1)$. While this receptive field may not cover the whole graph, we assume that the node and edge features can be evaluated locally.  

It has been shown \cite{Arvind2020} that MPNN cannot capture graph substructures other than forests of stars. To alleviate this issue, we further embody the graph topology into the MPNNs by extending the node representation by a set of skeleton-related features. Similarly to \cite{Bouritsas2020}, we complement the node hidden states at the initial step $\rvh_i^{(0)}$ of all generators and critics by the node degree $d_i$ and the number of $k$-cycles (cycles of length $k$ the node is part of) extracted from the graph skeleton.

\subsection{Graph representations}\label{sec:representation}

An important property of graph representation $\gG$ are their non-uniqueness. 
There are in general $n!$ possible permutations $\pi$ determining the ordering of the nodes.
The particular choice of the representation ordering from the complete set $\Pi = \{ \pi_i \}_{i=1}^{n!}$ is therefore another source of stochasticity so that
\begin{align}\label{eq:permuations}
p(\gG) = p\big( \bigcup_{\pi \in \Pi} \pi(\gG) \big) = \sum_{\pi \in \Pi} p\big( \pi(\gG) \big) \enspace .
\end{align}

From equation \eqref{eq:permuations} we observe that when relying on the ordered graph representations $\gG$, one shall in principle model the complete set of distributions $p(\pi(\gG))$ for all the $n!$ permutations $\pi$ to capture the unordered-set graph distribution $p(\gG)$.
However, operating over the individual permutations $\pi(\gG)$ would lead to inefficient use of the model capacity.
As an alternative, previous methods often fallback to operating over a single representation of each graph using some heuristic to fix the canonical ordering (such as the minimum depth-first-search \cite{goyal_graphgen_2020}, the unique breadth-first-search \cite{you_graphrnn_2018}  or the further sequentiallized domain driven SMILES \cite{daylight_chemical_information_systems_inc_daylight_2011}).
In result, the model has to learn not only the distribution of the graphs but also the ordering heuristic which is again inefficient from the perspective of generating the unordered graph sets $\gG$.

A fundamental property of GrannGAN is that it is completely insensitive to the ordering $\pi(\gG)$ that can in result be chosen arbitrarily.
This permuation equivariance\footnote{A function $f$ is equivariant with respect to permutation $\pi$ if $f(\pi(\vx)) = \pi(f(\vx)).$} of the message passing operations is pivotal for our method as it solves the problem of the graph representation non-uniqueness while allowing for efficient use of the model capacity.

\section{Related work}\label{sec:related}

GrannGAN is closely related to the broad category of graph generative models, an area which has attracted significant attention of the research community resulting in a flurry of papers in the last several years. A systematic review of the major advancements in the field can be found for example in the excellent survey \cite{faez_deep_2021}.

Unlike  previous work on that field, our method focuses on the particular problem of generating node and edge features conditionally over a given graph skeleton.
To the best of our knowledge, we are the first to investigate such structural-based conditional graph feature generation.

The closest to our settings are methods modelling the distribution of annotated graphs. These can be categorized into two large families: the ones generating graphs sequentially and those generating in the whole graph in one go.

\subsection{Sequential graph generation}
The first are methods generating the graphs sequentially, starting from small structures to which they gradually connect new graph components (nodes, edges, or complete sub-structures).
Most of these focus on generating chemical molecules and adopt various measures to improve the performance on this particular generative problem.
CharacterVAE \cite{gomez-bombarelli_automatic_2018} and GrammarVAE \cite{kusner_grammar_2017} rely directly on sequential SMILES representation of the molecules.
JT-VAE \cite{jin_junction_2018} operates over hand-crafted vocabulary of chemically valid sub-structures. 
MolecularRNN \cite{popova_molecularrnn_2019} and GraphAF \cite{shi_graphaf_2019} generate the graphs by successive node and edge sampling steps
and ensure validity of the generated molecules by valency checking and the possibility for resampling at each of these steps.

Thanks to these domain-motivated components, the sequential methods often achieve excellent performance on molecular graphs datasets.
However, those directly relying on chemistry-specific data representations (e.g SMILES) cannot be easily extended to beyond the molecular problems.
On the other hand, methods not using the SMILES representation need to establish their own heuristic for the sequential traversal of the graph such as the breadth-first-search in GraphRNN \cite{you_graphrnn_2018} which in turn needs to be learned by the generative model together with the graph data distribution.  
As any other sequential models, the graph autoregressive generators need to be particularly careful about capturing long-term dependencies, and suffer from slow sampling process.

Different from the previous models NetGAN \cite{netgan} uses GANs to learn the distribution of random walks over one big graph.

\subsection{Annotated graph generation in one go}
The other large family of models, into which GrannGAN can also be related, are those generating the graphs in one go, that is all the nodes and edges and their features together.
GraphVAE \cite{simonovsky_graphvae_2018} proposes to sidestep the problem of graph linearization characteristic for sequential models by relying on the variational auto-encoder framework \cite{DBLP:journals/corr/KingmaW13}.
Due to modelling the graph in the ordered representation of annotation and adjacency matrix, it needs to employ an expensive (though inexact) graph-matching procedure in the training loss calculation, which significantly hampers its scalability.
The flow-based graphNVP \cite{madhawa_graphnvp_2019} adopts the coupling strategy of \cite{DBLP:conf/iclr/DinhSB17} applying it to the rows of the graph annotation and adjacency matrices and thus preserving the ordering of the nodes and edges through the flow. 
GraphNVP is the only model for annotated graph using permutation equivarient generative function. 
In general, annotated graph generation in one go is still an open field of research.
Our contribution can be seen as a proposition to tackle this challenging issue.

Misc-GAN \cite{miscGan} aims to translate graphs from a source-domain graph into a target-domain graph using a multiscale cycle-GAN. It operates on unannotated graphs.  

The closest to our GrannGAN is the MolGAN \cite{de_cao_molgan_2018} model, which uses GAN for molecule generation. Unlike in GrannGAN, the MolGAN generator is a feed-forward network sampling the ordered annotation and adjacency matrices of the graph representation. The authors of MolGAN observe that the model tends to suffer from mode collapse resulting in insufficient variation in the generated samples.  We suppose that this issue comes from the loss of capacity by learning to generate various permutations of the same graph and from an additional Reinforcement Learning module encouraging the generator to produce graphs with some specific properties. We never experienced mode collapse during our experiments. As we show it in the experiment section, our model presents excellent uniqueness and novelty rates. 
We compare to MolGAN and other methods discussed here in our experiments in section \ref{sec:experiments}.

\section{Experiments}\label{sec:experiments}

In this section, we present experiments using our model as a graph generative model. 
There currently exists no other models conditioning on the graph skeleton and, therefore, no other method to directly compare with. 
Instead, we use generative models presented in the previous section as baselines. 
However, we underline that these models do not have the same purpose. 
In the second part of this section, we also present the results of the conditional generation by fixing the skeleton. 

We evaluate our method in a set of experiments over three datasets containing (node- and edge-)  annotated graphs\footnote{Unfortunately, we did not found other public dataset with enough node- and edge-annotated graphs.}. 
Two of the datasets, QM9 \cite{ramakrishnan_quantum_2014} and ZINC \cite{sterling_zinc_2015}, are from the chemical domain that is frequently used as a test bed for annotated graph modeling.
The third is the fingerprint dataset \cite{Riesen2008} included in the TUDataset collection \cite{Morris+2020}.

As it has recently been discussed, for example in \cite{obray_evaluation_2021} and \cite{thompson_evaluation_2021}, evaluating generative models of graph structures is particularly challenging due to the impracticality of visual or perceptual comparisons of the data examples. The comparison with existing graph generative models is therefore difficult and can provide only crude indication of the method capabilities.

For the chemical datasets the most common evaluation metrics are the following
\begin{itemize}
\item \emph{validity} is the proportion of chemically valid molecules in the total generated examples and measures the ability of the method to understand the chemical constraints differentiating general graphs from chemically valid molecules
\item \emph{uniqueness} is the proportion of unique samples within the \emph{valid} generated samples and measures the variability of the generated data
\item \emph{novelty} is the proportion of examples within the \emph{valid} and \emph{unique} set that do not exist in the training dataset and measures the ability of the method to go beyond data memorization
\end{itemize}
We complement these by an \textbf{overall} score of \emph{valid-unique-novel} molecules calculated simply as the product of the three above metrics and measuring the overall quality of the generated examples as a proportion of samples with all three of these desirable properties in the set of generated data. So, the overall score gives the rate of valid molecules that are neither in the dataset nor already generated. Note that in most cases the overall score is the metric of interest.

While these evaluation metrics may be useful indicators for chemical downstream tasks, they do not provide any measure of the distance between the distribution of reference and the generated distributions.
Various indicators calling on the framework of maximum mean discrepancy (MMD) \cite{gretton_kernel_2012} have been proposed in the literature.
These are used very inconsistently and, as demonstrated in \cite{obray_evaluation_2021}, they are highly sensitive to the specific choice of the graph statistics, the kernel and the hyperparameters used for the MMD calculation.
Instead of calculating the MMDs on graphs statistics, we report directly the distance between the generated feature distributions, using the Jenson-Shannon Distance for each feature. We show that with respect to these metrics, our model outperforms by far GraphAF, model considered as the state-of-the-art. 

\subsection{QM9}\label{sec:qm9}

QM9 is one of the most commonly used datasets for testing models for annotated graph generation.
It consists of $\sim$134k stable organic molecules with up to 9 atoms of four types.
The 4 atom types and 4 bond types are encoded as one-hot vectors in the node feature descriptions $\rvv$.

In table \ref{tab:qm9} we present an overview of the generative performance of our GrannGAN method in comparison with a set of well-established graph generative methods: MolGAN \cite{de_cao_molgan_2018}, GraphVAE \cite{simonovsky_graphvae_2018}, GraphNVP \cite{madhawa_graphnvp_2019}, 
GraphAF \cite{shi_graphaf_2019}, CharacterVAE \cite{gomez-bombarelli_automatic_2018}, GrammarVAE \cite{kusner_grammar_2017}, and JT-VAE \cite{jin_junction_2018}.

The GrannGAN results are calculated from 1000 new data examples generated by conditioning on skeletons randomly sampled from the training data.
The results of the baseline methods are those reported by the authors in the original papers\footnote{CharacterVAE and GrammarVAE results (not reported in the original papers) are taken over from the method replications in GraphNVP.}.
Despite the frailty of such comparisons due to inconsistencies in the original experimental protocols (for example, some of the methods, such as GraphAF, work over the \emph{kekulized} versions of the molecules reducing the number of edge categories to three), our GrannGAN achieves excellent results. It outperforms all one-go generative models and is on par with the best auto-regressive model. 

\setlength\tabcolsep{5pt} 
\begin{table}[htp]
\begin{center}
        \caption{QM9: performance comparison}\label{tab:qm9}
        \begin{tabular}{ c  l | c c c | c }
        \hline
        & \textbf{Model} &  \textbf{valid} & \textbf{unique} & \textbf{novel} & \textbf{overall}   \\
        \hline \hline
        \rowcolor{lgray}
        \multirow{5}{*}{\begin{sideways} no chemistry \end{sideways}} & GrannGAN & 82.5 & 99.9 & 64.9 & 53.4 \\
        & MolGAN (wo. RL)  & 87.7 & 2.9 & 97.7 & 2.5 \\ 
        & GraphVAE  & 55.7 & 76.0 & 61.6 & 26.1 \\ 
        & GraphNVP & 83.1 & 58.2 & 99.2 &  46.8 \\ 
        & GraphAF (wo. validity) & 67.0 & 94.5 & 88.8 & 56.3 \\ 
        \hline \hline
        \multirow{5}{*}{\begin{sideways} \gtext{ chemistry} \end{sideways}} & \gtext{MolGAN-RL} & \gtext{99.8} & \gtext{2.3} & \gtext{97.9} & \gtext{2.2} \\ 
        & \gtext{CharacterVAE} & \gtext{10.3} & \gtext{90.0} & \gtext{67.5} & \gtext{11.9} \\ 
        & \gtext{GrammarVAE} & \gtext{60.2} & \gtext{80.9} & \gtext{9.3} & \gtext{11.9} \\ 
        & \gtext{GraphAF} & \gtext{100} & \gtext{94.5} & \gtext{88.8} & \gtext{83.9} \\ 
        \hline
        \end{tabular}
\end{center}
\end{table}

As explained above, the metrics in the table are very much domain specific.
Methods in the lower part of table \ref{tab:qm9} expressly focus on these
introducing into the models various expert-designed modules promoting chemical validity of the generated molecular graphs (e.g., valid sub-substructures vocabulary, valency constraints, rejection sampling, etc.).
While these improve the targeted metrics in table \ref{tab:qm9}, they also bias the generative process and, therefore, the model distribution towards the validity metric in exchange for the distribution approximation goal.
We list these methods in table \ref{tab:qm9} for completeness despite them not being really comparable to our approach which focuses on the distribution approximation of general graphs.

\begin{figure}[htp]
\centering
\includegraphics[width=0.3\textwidth]{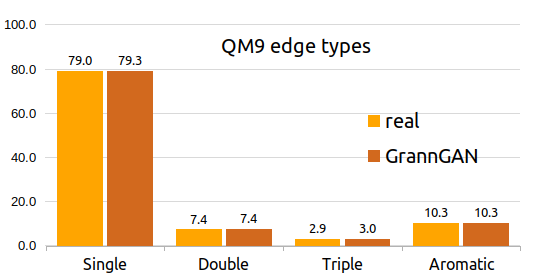}
\includegraphics[width=0.3\textwidth]{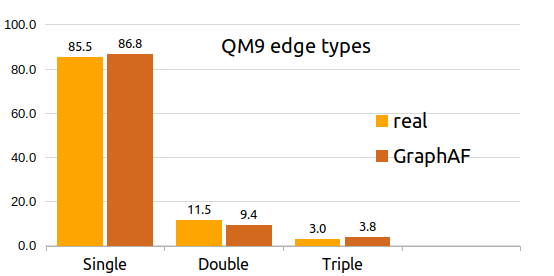}\\
\includegraphics[width=0.3\textwidth]{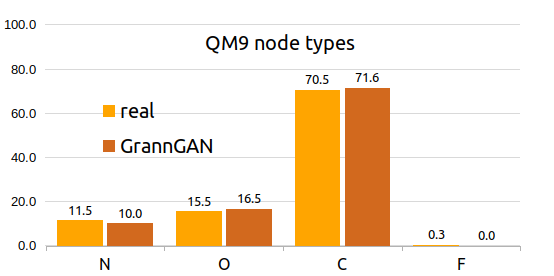}
\includegraphics[width=0.3\textwidth]{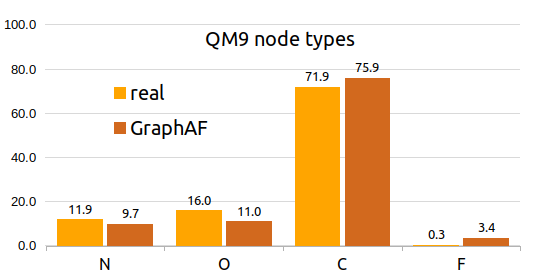}
\caption{QM9: empirical distributions of real and generated features  (samples of 1000 graphs)}
\label{fig:qm9}
\end{figure}

\setlength\tabcolsep{5pt} 
\begin{table}[htp]
\begin{center}
        \caption{Jenson-Shannon distances between real and generated distributions for node and edge features of the QM9 and ZINC datasets (the lower the better)}\label{tab:JSmolecule}
        \begin{tabular}{ l | c c c c}
        \hline
        \textbf{Model} &  \textbf{nodes - QM9} & \textbf{edges - QM9} &  \textbf{nodes - ZINC} & \textbf{edges - ZINC}    \\
        \hline \hline
        \rowcolor{lgray}
        GrannGAN & $1.87 \cdot 10^{-3}$ & $ 0.04 \cdot 10^{-4}$ & $1.50 \cdot 10^{-3}$ & $0.60 \cdot 10^{-3}$\\
        GraphAF  & $15.80 \cdot 10^{-3}$ & $11.5 \cdot 10^{-4}$ & $15.5 \cdot 10^{-3}$ & $3.39 \cdot 10^{-3}$ \\
        
        \hline
        \end{tabular}
\end{center}
\end{table}

Figure \ref{fig:qm9} documents the ability of the GrannGAN method to learn the feature distribution by comparing the empirical real data distribution with the distribution of the generated examples.
We have replicated the results of the GraphAF method, which is the most competitive with GrannGAN according to table \ref{tab:qm9}), to extract similar statistics.
GrannGAN matches the feature distributions of the reference dataset almost exactly while we observe a less closer match in the GraphAF\footnote{GraphAF uses somewhat different training set than we. For example, by kekulizing the molecules it removes all aromatic edges and thus reduces the number of edge types to 3. This is true also for the ZINC dataset.} generated data. 
This observation is numerically confirmed in table \ref{tab:JSmolecule} listing the Jensen-Shannon distances (JSD) between the real and generated data samples, where GrannGAN is systematically better by one order of magnitude.

\subsection{ZINC}\label{sec:zinc}

We use here the ZINC250k version used previously for generative modeling evaluations.
It consists of 250k randomly selected molecules from the complete ZINC set with up to 38 atoms of 9 types and 4 bond types.

Table \ref{tab:zinc} summarizes the generative performance of GrannGAN over 1000 newly synthesized examples compared to results reported in the literature. 
MolGAN and GraphVAE did not experiment (or did not provide the results) on this more challenging dataset of larger graphs.
GrannGAN scales to this dataset easily and performs competitively with respect to the chemically motivated metrics (keeping in mind the difficulties of such comparisons related to differences in experimental protocols and evaluation procedures).

\begin{figure}[htp]
\centering
\includegraphics[width=0.3\textwidth]{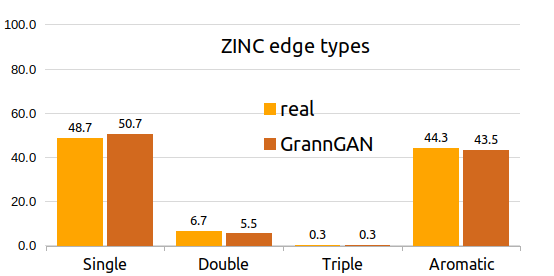}
\includegraphics[width=0.3\textwidth]{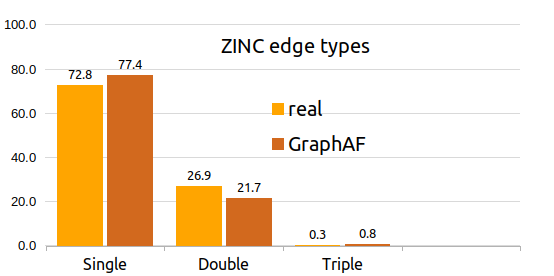}
\\
\includegraphics[width=0.3\textwidth]{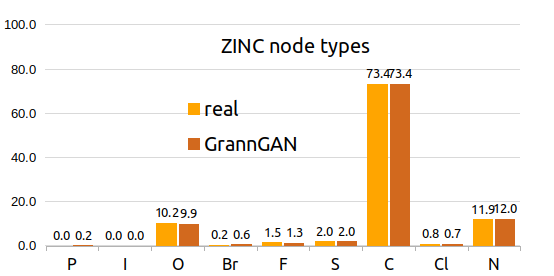}
\includegraphics[width=0.3\textwidth]{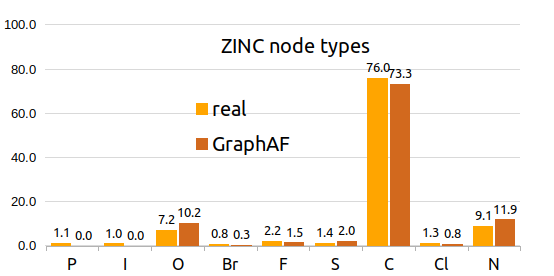}
\caption{ZINC: empirical distributions of real and generated features (samples of 1000 graphs)}
\label{fig:zinc}
\end{figure}

Figure \ref{fig:zinc} documents the excellent performance of GrannGAN in learning the feature distributions, matching it closely for both the node and edge features.
The differences in the true and GraphAF generated data distributions are more pronounced as is also clear from the numerical evaluation via the JSD presented in table \ref{tab:JSmolecule}.

\setlength\tabcolsep{5pt} 
\begin{table}[htp]
\begin{center}
        \caption{ZINC: performance comparison}\label{tab:zinc}
        \begin{tabular}{ c  l | c c c | c }
        \hline
        & \textbf{Model} &  \textbf{valid} & \textbf{unique} & \textbf{novel} & \textbf{overall}   \\
        \hline \hline
        \rowcolor{lgray}
        \multirow{5}{*}{\begin{sideways} no chemistry \end{sideways}} & GrannGAN & 56.5 & 100 & 100 & 56.5 \\
        & MolGAN (wo. RL)  & --- & --- & --- & --- \\
        & GraphVAE  & 13.5 & --- & --- & --- \\ 
        & GraphNVP & 42.6 & 94.8 & 100 &  40.4 \\ 
        & GraphAF (wo. validity) & 68.0 & 99.1 & 100 & 67.3 \\
        \hline \hline
        \multirow{5}{*}{\begin{sideways} \gtext{chemistry} \end{sideways}} & \gtext{MolGAN-RL} & \gtext{---} & \gtext{---} & \gtext{---} & \gtext{---} \\
        & \gtext{CharacterVAE} & \gtext{7.2} & \gtext{9.0} & \gtext{100} & \gtext{0.6} \\ 
        & \gtext{GrammarVAE} & \gtext{0.7} & \gtext{67.5} & \gtext{100} & \gtext{0.5} \\ 
        & \gtext{JT-VAE} & \gtext{100} & \gtext{100} & \gtext{100} & \gtext{100} \\ 
        & \gtext{GraphAF} & \gtext{100} & \gtext{99.1} & \gtext{100} & \gtext{99.1} \\ 
        \hline
        \end{tabular}
\end{center}
\end{table}


\subsection{Fingerprints}\label{sec:finger}

Fingerprint is a dataset of graphs representing the skeletonized regions of interest in human fingerprints \cite{Riesen2008}.
The graphs consist of up to 26 nodes. 
Both nodes and edges are described by 2 continuous features related to their position and orientation in the 2d fingerprint image.
We have linearly re-scaled both node and edge features between -1 and 1. 

We use the non-annotated graphs as the skeletons for new feature generations.
Each row displays graphs arranged according to the new features generated by GrannGAN.

\begin{figure}[htp]
\centering
\includegraphics[width=0.25\textwidth]{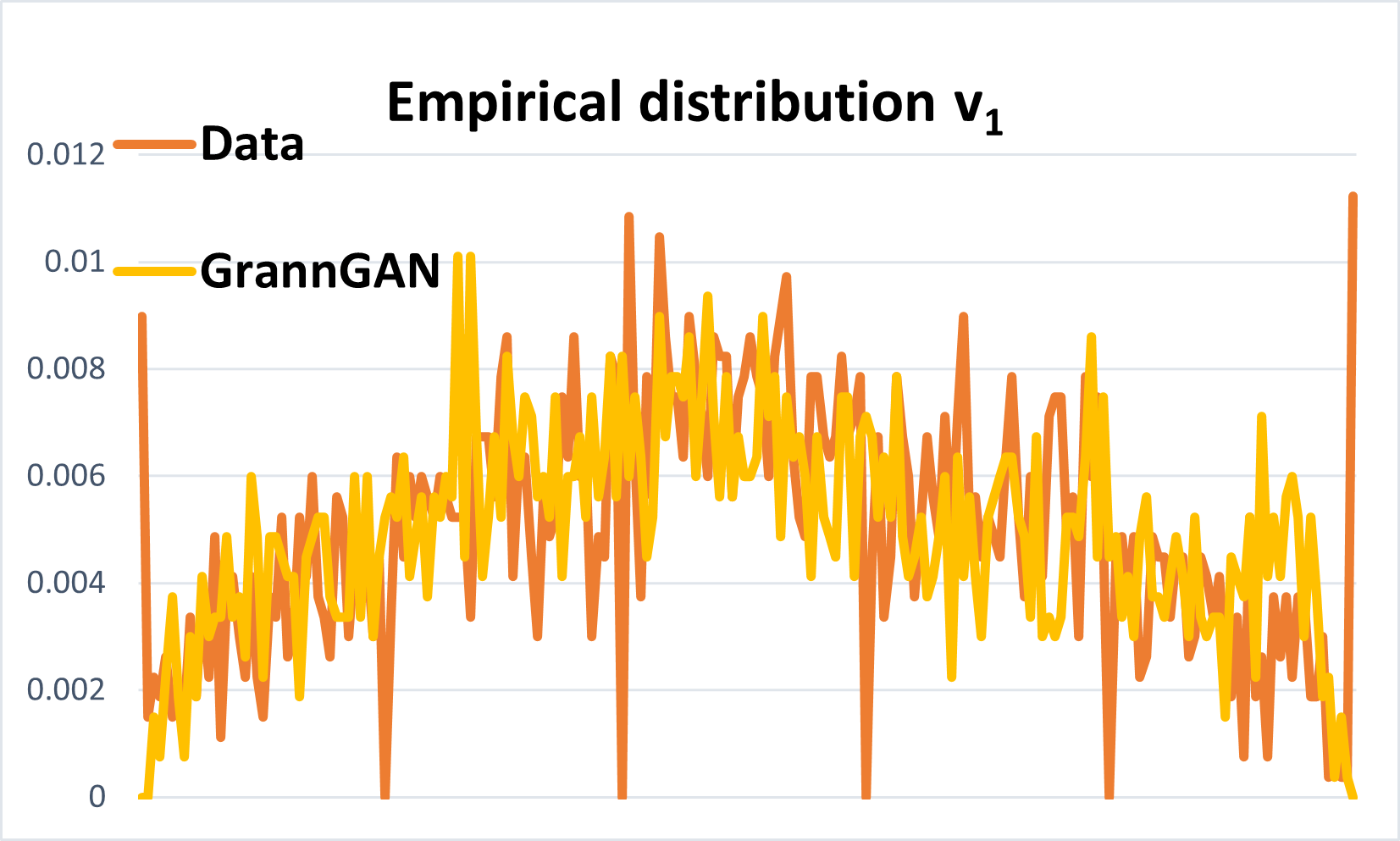}
\includegraphics[width=0.25\textwidth]{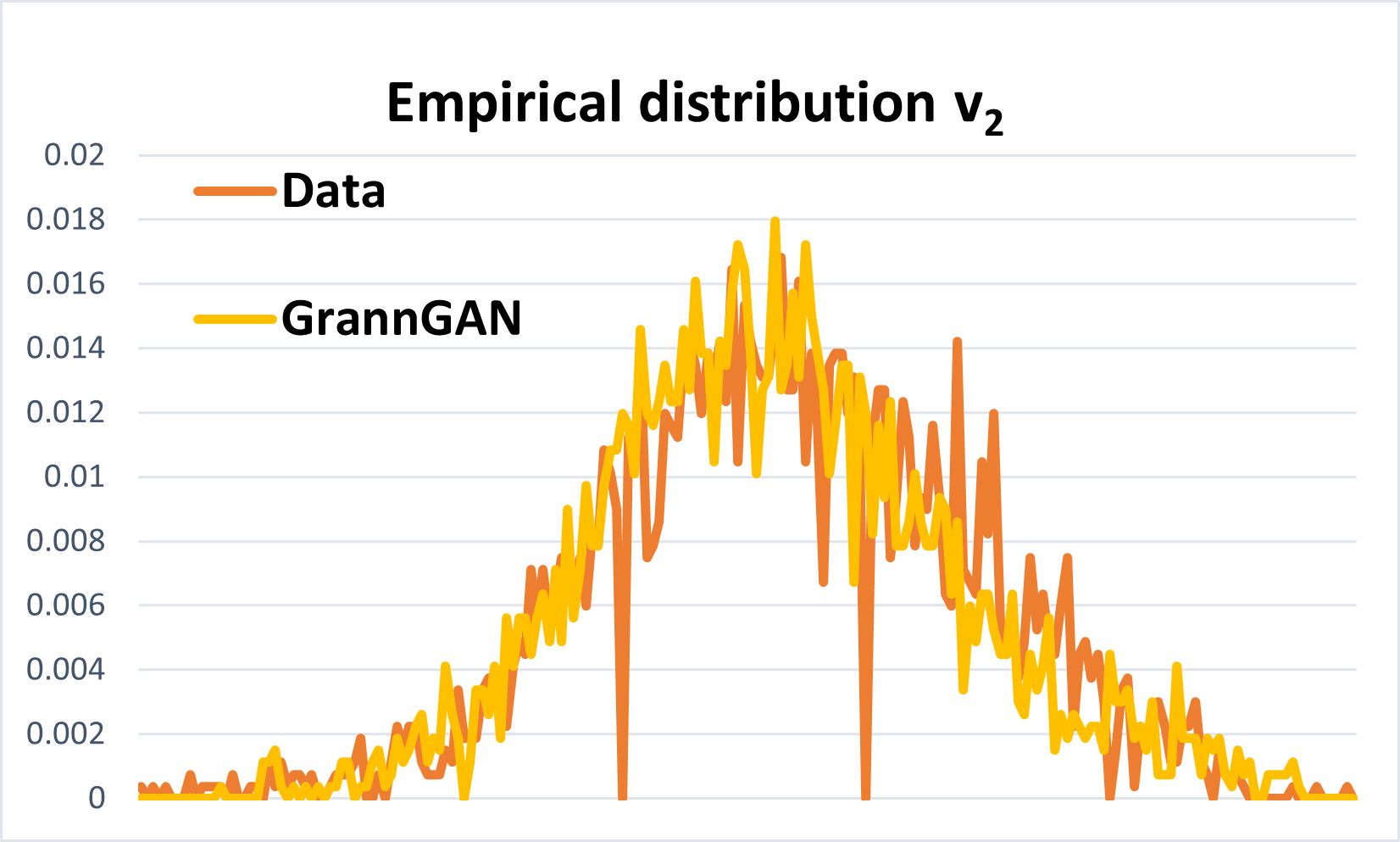}
\includegraphics[width=0.25\textwidth]{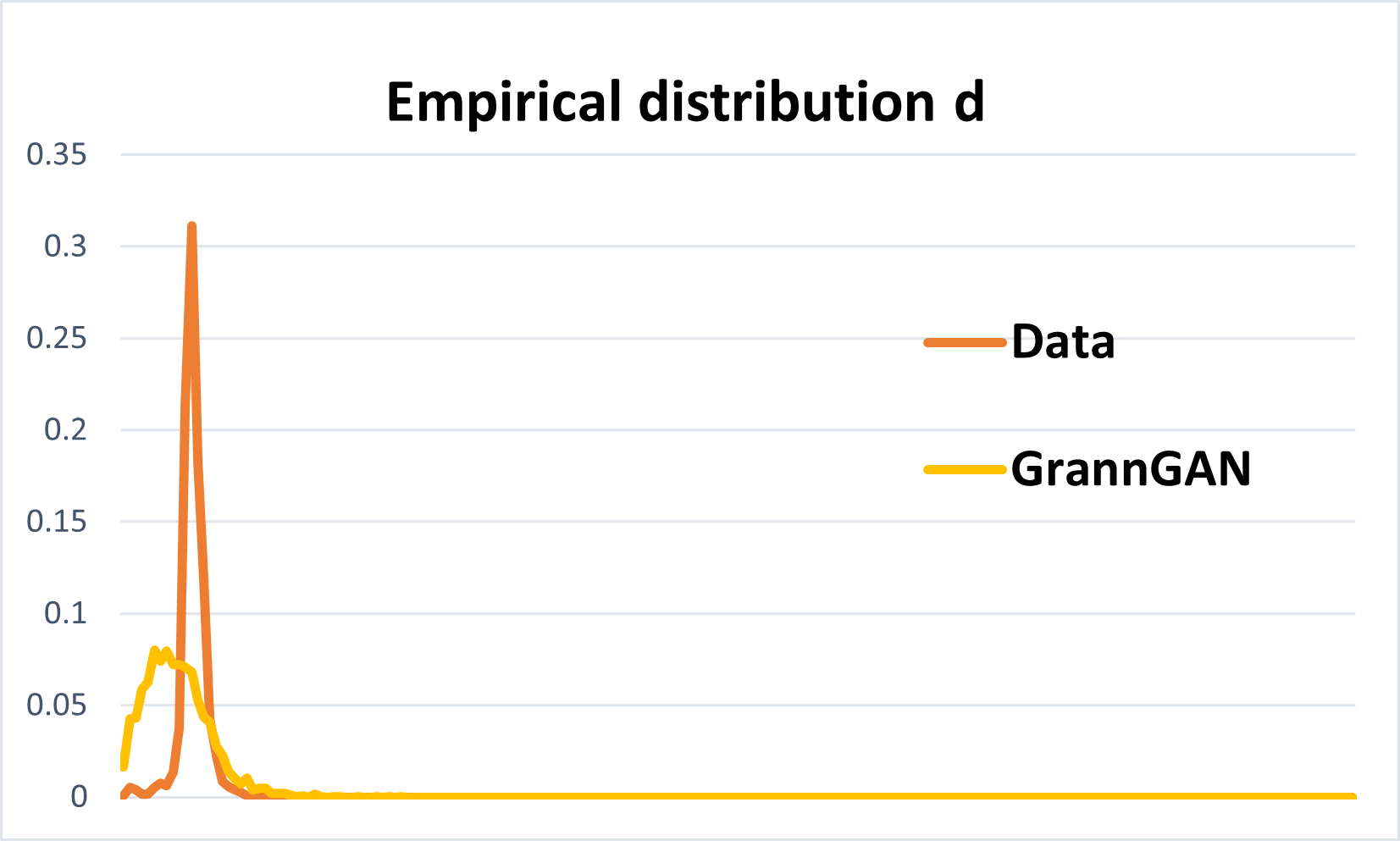}
\caption{Fingerprints: empirical distributions $\rvv_1$, $\rvv_2$ and $\rvd$ sampled from our GrannGAN and from the dataset  (samples of 1000 graphs)} 
\label{fig:FP_node}
\end{figure}

\begin{figure}[htp]
\centering
\includegraphics[width=0.3\textwidth]{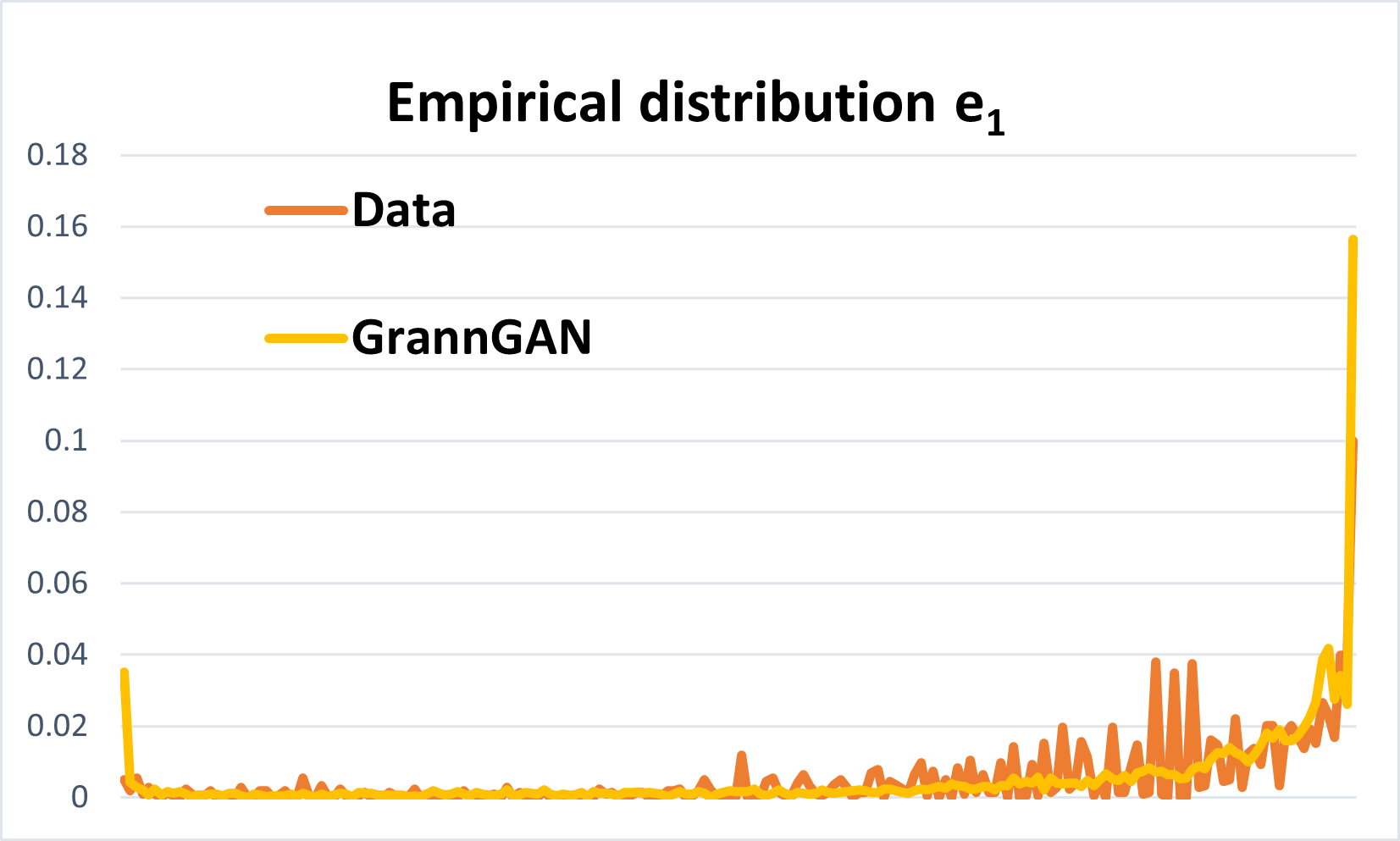}
\includegraphics[width=0.3\textwidth]{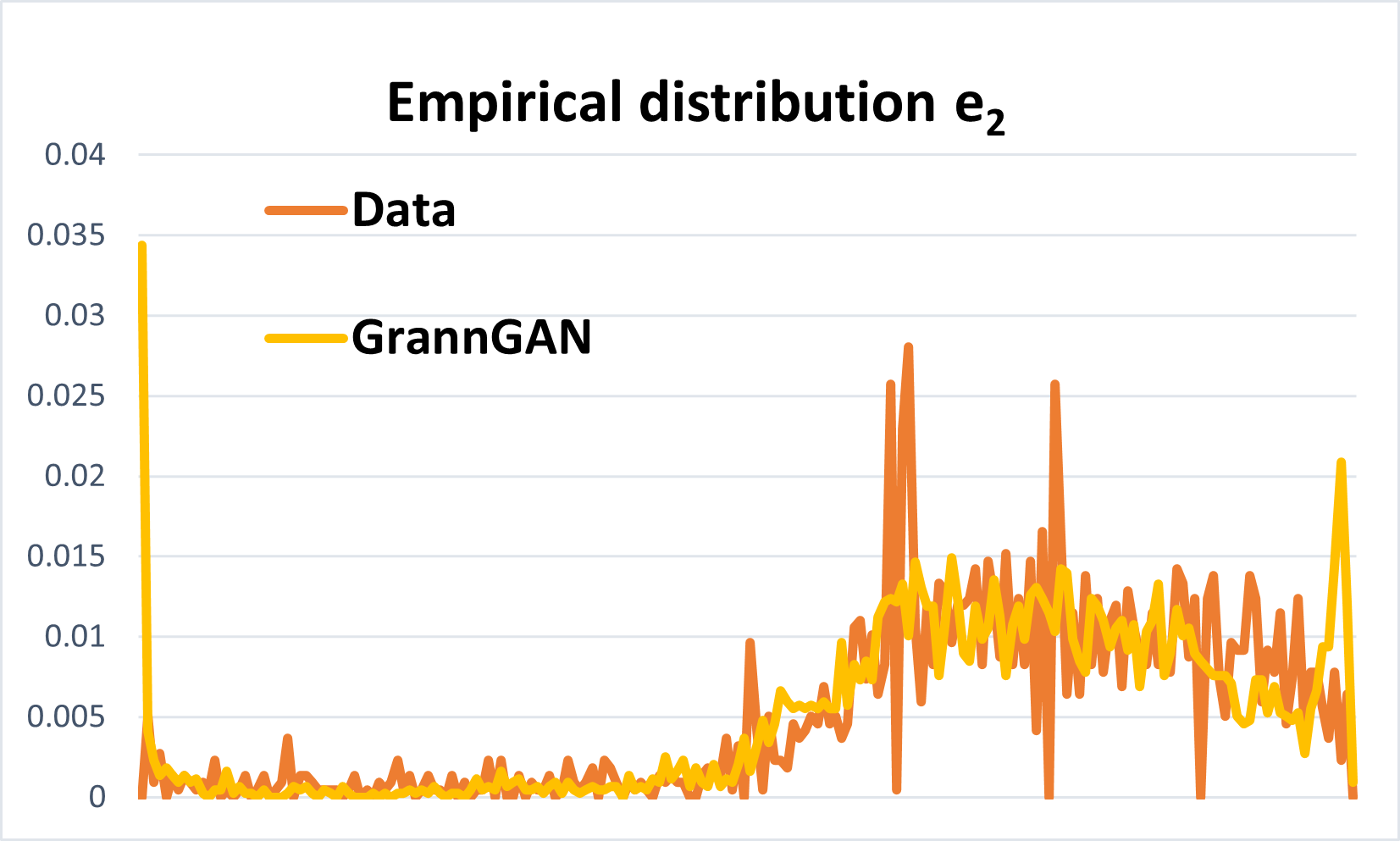}

\caption{Fingerprints: empirical distributions $\rve_1$ and $\rve_2$ sampled from our GrannGAN and from the dataset  (samples of 1000 graphs)}
\label{fig:FP_edge}
\end{figure}

In table \ref{tab:fingerprint} we list the JS distance (with the logarithm in base 2) between the empirical distributions of the training data and newly generated features starting from 500 randomly selected skeletons.
These have been calculated by discretizing the features into 200 even sized bins.
The columns in the table correspond to the two node $\rvv$ and edge $\rve$ features respectively.
Following the same methodology, we also report the JS distance in the distribution of the euclidean distances between connected nodes $\rvd$. 
Unlike the feature distribution, this statistic consider the relation between node features. 
In figures \ref{fig:FP_node} and \ref{fig:FP_edge}, we show the node and edge features distribution for the data and newly generated examples from our model as well as the distribution of distances between node $\rvd$\footnote{We assume that the oscillation of the curve is just an artifact coming from the relatively small data points by bin}.

\setlength\tabcolsep{5pt} 
\begin{table}[htp]
\begin{center}
        \caption{Jenson-Shannon distances between real and generated distributions for node and edge features of the Fingerprint dataset (the lower the better)}\label{tab:fingerprint}
        \begin{tabular}{c c c c c c}
        \hline
        &\textbf{v1} &  \textbf{v2} & \textbf{d} & \textbf{e1} & \textbf{e2} \\
        \hline
        \textbf{GrannGAN} & $5.02 \cdot 10^{-2}$ &  $4.55 \cdot 10^{-2}$ & $32.84 \cdot 10^{-2}$ & $16.78 \cdot 10^{-2}$ & $9.74 \cdot 10^{-2}$ \\
        \hline
        \textbf{Data samples} & $1.13 \cdot 10^{-2}$ &  $1.27 \cdot 10^{-2}$ & $1.49 \cdot 10^{-2}$ & $1.37 \cdot 10^{-2}$ & $1.07 \cdot 10^{-2}$ \\
        \hline
        \end{tabular}
\end{center}
\end{table}

Again, our model shows excellent performance in closely matching the node and edge features and their relations.

These experiments show the quality of our GrannGAN for the generative task. 
It outperforms all previous one-go generative models. 
The sequential GraphAF is the only model presenting slightly better results using the molecular metrics. 
However, we show that our model is much better at capturing the feature distributions. 

\subsection{Conditional generation}
In this section, we show that our model can be used for conditional generation by fixing the skeleton and still producing a high level of diversity.

We want to guarantee that the sample diversity does not come mainly from the skeleton. To test the generative capacity of our GrannGAN, we randomly sampled 100 skeletons from the ZINC dataset. We generate 1000 new instances conditioned on each sampled skeleton. We compute the average rate of uniqueness, i.e. the rate of unique instances over all the valid molecules.

Table \ref{tab:condgen} reports the results of the experiment. 
As expected, the validity rate is similar to the one by sampling a new skeleton for each instance. 
Even by fixing the skeleton, we reached a rate of 92.5\% unique instances on average.
On average, the conditional generation of 1000 instances over the same skeleton produces 527.0 unique molecules.
So, we show that the performance of our model does not depend critically on the sampling of multiple skeletons and can be properly used as a conditional generative model. 

\setlength\tabcolsep{5pt} 
\begin{table}[htp]
\begin{center}
        \caption{Conditional generation of 1000 instances from the same skeleton (average)}\label{tab:condgen}
        \begin{tabular}{c | c c c}
        \hline
        \textbf{Model} &  \textbf{Validity (\%)} &  \textbf{Uniqueness (\%)} & \textbf{Valid and unique (\textperthousand)} \\
        \hline
        \textbf{GrannGAN} & $57.0$ &  $92.5$ & $527.0$  \\
        
        \hline
        \end{tabular}
\end{center}
\end{table}

We have demonstrated that the GrannGAN is competitive with state-of-the-art model for graph generation even though it is first meant for conditional generation. 
GraphAF is slightly better on the molecular metric, but much worst at matching the feature distribution. 
In addition, our model generates instances in one go and is therefore much faster during inference. 
More fundamentally, our model preserves its quality while generating new instances conditioned on the skeleton, a task that no other model can do. 

\section{Conclusions}\label{sec:conclude}

We have presented here a new method, GrannGAN, for implicit distribution learning and generating features of nodes and edges conditioned on graph skeletons.
The promising results of the method, as documented in our experiments, indicate that the method can learn the complex high-dimensional distributions and generate new data examples with features coherent with the underlying graph structure.
When the generation of new skeletons is not crucial, GrannGAN can be used as a full generative model by sampling the skeleton from the data.
It is also the first method for graph generation conditioned on the skeleton. 

These favorable results provide a solid starting point for multiple possible extensions in future work.
For example, in some applications (e.g. bio-chemistry) a reasonable starting point for the feature generations may be a partly annotated graph (in place of a completely non-annotated graph).
Our model, which is already conditional, could be easily adapted to this setting by re-using the ideas of our second edge-annotation phase.
Another interesting extension is that of property optimization.
An additional conditioning variable could be included in the generators and critics to guide the model in generating property-conditioned features.

\acks{The contribution of Magda Gregorova has been supported by High-Tech Agenda Bayern.}

\bibliography{references}

\end{document}